# Bridging Geometry-Coherent Text-to-3D Generation with Multi-View Diffusion Priors and Gaussian Splatting

Feng Yang[a], Wenliang Qian[a], Wangmeng Zuo[b], Hui Li[a,b,*]

[a]*Key Lab of Smart Prevention and Mitigation of Civil Engineering Disasters of the Ministry of Industry and Information Technology, Harbin Institute of Technology, Harbin 150090, China*
[b]*School of Computer Science and Technology, Harbin Institute of Technology, Harbin 150001, China*

**Abstract**

Score Distillation Sampling (SDS) leverages pretrained 2D diffusion models to advance text-to-3D generation but neglects multi-view correlations, being prone to geometric inconsistencies and multi-face artifacts in the generated 3D content. In this work, we propose Coupled Score Distillation (CSD), a framework that couples multi-view joint distribution priors to ensure geometrically consistent 3D generation while enabling the stable and direct optimization of 3D Gaussian Splatting. Specifically, by reformulating the optimization as a multi-view joint optimization problem, we derive an effective optimization rule that effectively couples multi-view priors to guide optimization across different viewpoints while preserving the diversity of generated 3D assets. Additionally, we propose a framework that directly optimizes 3D Gaussian Splatting (3D-GS) with random initialization to generate geometrically consistent 3D content. We further employ a deformable tetrahedral grid, initialized from 3D-GS and refined through CSD, to produce high-quality, refined meshes. Quantitative and qualitative experimental results demonstrate the efficiency and competitive quality of our approach.

## 1. Introduction

3D content creation is crucial across various fields, including gaming, virtual reality, industrial design, architecture creation, and environment simulation. Despite the immense potential of 3D creation, traditional methods, such as manual modeling using software like Blender or Maya, of producing high-quality 3D content are resource-intensive and require considerable expertise and time. Automating 3D content generation is essential for improving availability. Integrating AI into 3D object generation not only streamlines the modeling process but also addresses the creative limitations of human designers. While many open-source 3D datasets (Deitke et al., 2023, 2024) exist, they are insufficient for directly training 3D generative models with strong generation capabilities, especially when compared to the vast scale of billions of image-text pairs used for training 2D diffusion models. For instance, generative models

* Corresponding author.
  *E-mail address:* lihui@hit.edu.cn (H. Li)

(Cao et al., 2023; Jun & Nichol, 2023; Nichol et al., 2022) trained directly on 3D data face challenges in generating arbitrary objects from text prompts. Recently, DreamFusion (Poole et al., 2022) has addressed the limited 3D data problem by utilizing 2D priors from pretrained diffusion models (Rombach et al., 2022) to optimize learnable parametric 3D representations, such as NeRF (Mildenhall et al., 2020), from random viewpoints, enabling general text-to-3D generation without the need for 3D-specific training data. Subsequent works (R. Chen et al., 2023; C.-H. Lin et al., 2023; Tang, Ren, et al., 2024; Wang et al., 2023) have further improved generation quality and rendering speed through the different optimization strategies or by utilizing alternative 3D representations, such as 3D Gaussian Splatting (3D-GS) (Kerbl et al., 2023).

Although these optimization methods can generate high-quality 3D assets, they suffer from problems such as the unreasonable topology and Janus problem (C.-H. Lin et al., 2023; Wang et al., 2023), resulting in repeated and unreasonable content from different viewpoints. Previous approaches did efforts to fix the problem by utilizing view-dependent prompts (Nalisnick et al., 2019; Wang et al., 2023), 3D priors (Yi et al., 2024), or 3D loss (Z. Chen et al., 2024). However, the Janus problem may still exist during optimization as the view is sampled one by one (T. Wu et al., 2024), and the perspective bias in the pretrained 2D diffusion model remains (Huang et al., 2024; J. Liu et al., 2024). Moreover, learnable 3D representations are crucial in the optimization process when utilizing 2D diffusion models for guidance. Previous methods (C.-H. Lin et al., 2023; Poole et al., 2022; Wang et al., 2023) typically relied on NeRF as the underlying 3D representation. Although these approaches have improved generation quality, they are hindered by lengthy optimization times due to volumetric rendering, limiting their practicality in real-world applications. Recently, 3D-GS (Kerbl et al., 2023) has attracted attention in 3D reconstruction, mainly for its efficient rendering and competitive visual quality. However, a critical drawback of 3D-GS is that it is extremely sensitive to initialization, which frequently leads to unstable optimization processes. This instability not only hinders the convergence of the models but also compromises the consistency and reliability of the generated 3D representations, especially in complex or diverse scenarios. Incorporating pretrained 3D point cloud priors (Y. Lin et al., 2024; Yi et al., 2024) can enhance optimization stability, however, it constrains the diversity of generation.

Thus, achieving geometrically consistent, diverse, and general text-to-3D generation with 3D-GS remains a significant challenge. These limitations arise from two issues: (1) the lack of multi-view consistency during optimization, and (2) the instability of 3D-GS during the optimization process. To address the first issue, recent studies have introduced point cloud regularization loss during the optimization for 3D-GS (Z. Chen et al., 2024) and fine-tuned view-aware diffusion models (Z. Hu et al., 2024; Shi et al., 2023) using multi-view rendered images from available 3D datasets. Although these methods have effectively reduced repetitive content across viewpoints and improved the overall quality of generated 3D assets, the geometries remain constrained by the regularization shape, and the diversity is limited by the fine-tuned models. For the second issue, current research (Yi et al., 2024) focuses on improving the stability of 3D-GS by initializing with coarse 3D priors from off-the-shelf 3D models (Jun & Nichol, 2023) and further fine-tuning the 3D-GS for realistic text-to-3D generation. However, this reliance on 3D priors significantly constrains the diversity of the generated results, as the geometry is strongly influenced by the initial priors.

In this work, we present Coupled Score Distillation (CSD), a method that achieves geometrically consistent and diverse text-to-3D generation while ensuring stable optimization of 3D-GS from random initialization. Specifically, to address the Janus problem, we reformulate the optimization process as a multi-view joint optimization task and derive a novel gradient-based update rule grounded in the probability product rule. This rule effectively facilitates multi-view joint optimization, ensuring geometric consistency in the generated 3D assets while preserving diversity. To further enhance the stability and photorealism of 3D-GS optimization, we propose a framework that integrates a LoRA (Low-rank Adaptation) (E. J. Hu et al., 2021) of pretrained diffusion and a fine-tuned multi-view diffusion model (Shi et al., 2023) to ensure direct, stable, and realistic optimization of 3D-GS from random initialization. LoRA effectively adapts optimization directions to capture texture details by updating low-rank weights of the pretrained diffusion model. Meanwhile, the fine-tuned multi-view diffusion model leverages learned perspective correlations to stabilize the optimization process of 3D-GS. Furthermore, we refine a deformable tetrahedral grid initialized from the optimized 3D-GS using the proposed CSD, resulting in high-fidelity, photorealistic textured meshes, as illustrated in Figure 1.

In summary, our main contributions are:

- We propose a novel optimization-based text-to-3D generation method that addresses the multi-face problem through multi-view joint optimization, producing geometrically consistent 3D Gaussian Splatting and high-fidelity textured meshes.
- We derive a gradient-based optimization rule, termed Coupled Score Distillation, based on the multi-view joint optimization method, to ensure geometrically consistent 3D generation.
- We introduce a novel framework based on the optimization rule, combining a LoRA of pretrained diffusion model and a fine-tuned multi-view diffusion model, to enable stable and realistic optimization of 3D Gaussian Splatting from random initialization.

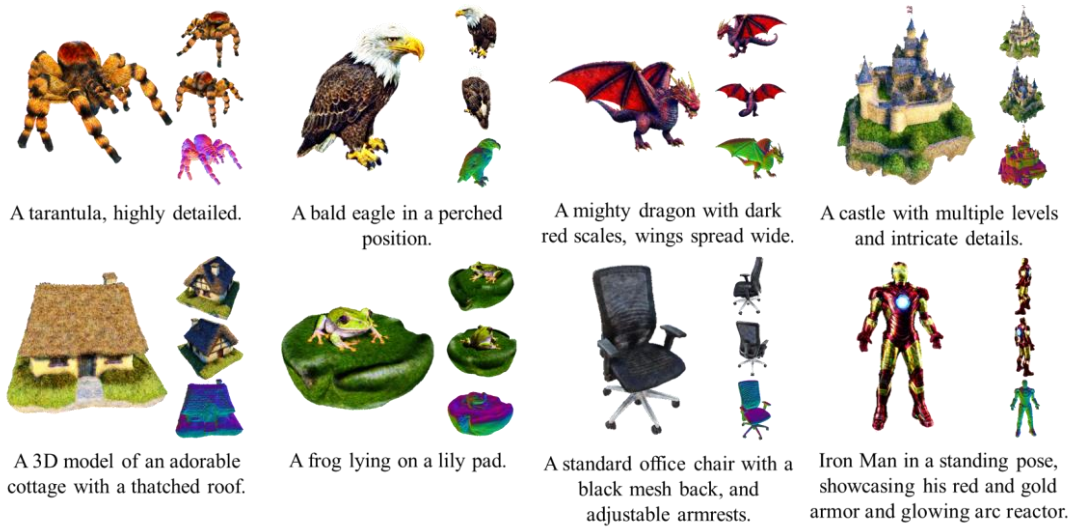

(a) Geometrically consistent and highly detailed 3D textured meshes generated by the proposed CSD. Each object is

depicted with three images captured from horizontal viewpoints spaced 120° apart, along with a corresponding mesh representation.

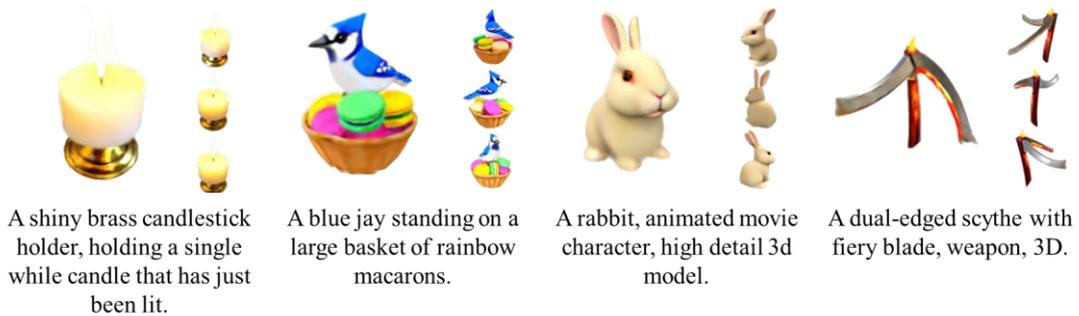

(b) Geometrically consistent and realistic 3D Gaussian Splatting generated by the proposed CSD. Each object is depicted with four images captured from horizontal viewpoints spaced 90° apart.

Fig. 1. Text-to-3D generation results using Coupled Score Distillation (CSD). The 3D Gaussian Splatting results are directly optimized using the proposed CSD, with positions randomly initialized within a unit sphere and colors set to grey. Mesh results are initialized from optimized 3D Gaussian Splatting and further refined with the proposed CSD.

## 2. Related work

**Text-to-3D generation.** Text-to-3D generation aims to produce 3D assets from text prompts and can be broadly categorized into three approaches. (i) Feed-forward methods (W. Li et al., 2024; S. Wu et al., 2024; Zhao et al., 2023), such as CLAY (Zhang et al., 2024) and TRELLIS (Xiang et al., 2024), typically employ a 3D VAE to encode shapes into a latent space, where a conditional latent diffusion model is trained to generate 3D objects. While these methods have shown improvements in geometry and diversity, they remain constrained by the limited availability of text-3D pairs. (ii) Text-to-image-to-3D methods (Y. Liu et al., 2024; Long et al., 2024), such as LGM (Hong et al., 2024), EfficientDreamer (Z. Hu et al., 2024), and Instant3D (J. Li et al., 2023), convert text prompts into multi-images and leverage sparse-view reconstruction to achieve 3D generation. While these methods have made remarkable advancements in multi-view consistency and generation quality, the absence of explicit geometric priors often leads to reduced geometric accuracy and surface detail inconsistencies (Guo et al., 2025). (iii) Optimization methods (Liang et al., 2024; Poole et al., 2022; Tang, Ren, et al., 2024, 2024) based on pretrained 2D diffusion model leverage score distillation sampling (SDS) to generate 3D assets. These methods optimize 3D representations by utilizing the priors learned from billions of text-image pairs. Although they typically involve longer generation times compared to the previous two paradigms, the integration of 3D representations and rich priors of the pretrained diffusion models enable the synthesis of more accurate, detailed, and realistic 3D assets. Moreover, they eliminate the reliance on text-3D data and can be used to generate high-quality training data for the previous two approaches.

**Optimization-based text-to-3D generation.** Diffusion models (Dhariwal & Nichol, 2021; Ho et al., 2020; Rombach et al., 2022; Song et al., 2020) have achieved remarkable results in the 2D generation domain. However, transitioning from 2D to 3D content generation remains challenging due to the lack of 3D data and limited computational resources. To address this, DreamFusion (Poole et al., 2022) first introduced SDS, a novel technique that leverages 2D diffusion priors as score functions to optimize learnable 3D representations without requiring 3D-specific data. Despite the success of SDS in text-to-3D, it still faces issues like over-saturation, simple geometry, and the multi-face Janus problem. To improve 3D fidelity, Magic3D (C.-H. Lin et al., 2023) introduced a coarse-to-fine framework for coarse NeRF (Mildenhall et al., 2020) optimization with fine DMTet (T. Shen et al., 2021) refinement. ProlificDreamer (Wang et al., 2023) modified the SDS objective with Variational Score Distillation (VSD), replacing noise $\epsilon$ with a trainable LoRA (E. J. Hu et al., 2021) of diffusion model to enable a common classifier-free guidance (CFG) (Ho & Salimans, 2022) weight. Fantasia3D (R. Chen et al., 2023) disentangled geometry and appearance for separate optimization, achieving improved 3D quality. DreamLCM (Zhong et al., 2024) incorporated the Latent Consistency Model (Luo et al., 2023) and improved guidance strategies to resolve the over-smoothing issue in SDS-based 3D generation, improving both quality and convergence speed. FlowDreamer (H. Li et al., 2024) combined the coupling and reversible properties of the rectified flow model (X. Liu et al., 2022) and introduced the Unique Couple Matching loss to enhance both generation quality and convergence speed. While these approaches enhance generated 3D content, they remain ineffective against the multi-face Janus problem. To address the Janus problem, recent studies (Nalisnick et al., 2019; Wang et al., 2023) utilized view-dependent prompts to incorporate viewpoint information during the optimization process. However, the Janus problem may still exist due to the one-by-one view sampling (T. Wu et al., 2024) and inherent perspective bias in the pretrained 2D diffusion models (Huang et al., 2024; J. Liu et al., 2024). Another recent study (Z. Chen et al., 2024) introduced additional point cloud regularization constraints, such as Point-E (Nichol et al., 2022), to enforce the geometrical consistency during the geometry optimization process. However, this method imposes strong regularization constraints, which heavily influence the generated geometries and limit the diversity of the results. DreamControl (Huang et al., 2024) proposed a two-stage 2D-lifting framework that refined coarse NeRF scenes into 3D self-priors and generated detailed objects through control-based score distillation. While effective, this approach struggles in extreme cases where priors exhibit similar appearances across different views, leading to Janus problem. Alternative methods, including MVDream (Shi et al., 2023) and Pi3d (Y.-T. Liu et al., 2024) fine-tuned view-aware diffusion models using rendered images from limited 3D datasets (Deitke et al., 2023, 2024) and subsequently optimize different learnable 3D representations based on SDS to achieve geometrically consistent text-to-3D generation. Despite their effectiveness, these fine-tuned models often suffer from overfitting, limiting their generalization to data beyond the training dataset. In this work, we frame the optimization process as a multi-view joint optimization problem and derive a formulation that couples multi-view joint distribution priors to jointly optimize multiple viewpoints at each optimization step, effectively addressing the multi-face Janus problem while maintaining generative diversity.

**Optimization-based 3D Generation with 3D Gaussian Splatting.** In optimization-based 3D generation, the choice of learnable parametric 3D representation plays a crucial role. NeRF (Mildenhall et al., 2020), utilizing volumetric rendering, has been widely adopted due to its strong performance in optimization-based 3D generation (C.-H. Lin et al., 2023; Poole et al., 2022; Wang et al., 2023; Xu et al., 2023; Zhu et al., 2024). However, the rendering process in NeRF is computationally inefficient due to its reliance on intensive volumetric sampling, which limits its practical applicability. Recently, 3D Gaussian Splatting (Kerbl et al., 2023) has emerged as a learnable representation capable of achieving rendering quality comparable to NeRF, but with the advantage of real-time rendering. This approach has been successfully applied in the optimization-based 3D generation field (Z. Chen et al., 2024; Q. Shen et al., 2025; Tang, Ren, et al., 2024; Yi et al., 2024). Although 3D Gaussians facilitate faster convergence, they are sensitive to initialization and prone to unstable optimization. To address this issue, LucidDreamer (Liang et al., 2024) utilized the pretrained text-to-point model (Nichol et al., 2022) for initialization and built upon DDIM (Song et al., 2020) to propose a deterministic diffusing trajectory for stable optimization, thereby avoiding perturbations in SDS-based methods. GaussianDreamer (Yi et al., 2024) and GSGEN (Z. Chen et al., 2024) utilized pretrained 3D point cloud generation models to provide priors, ensuring stable optimization and alleviating the Janus problem. However, relying on such pretrained models constrains the diversity of the generated results, as the geometry is heavily influenced by the initial 3D point cloud priors. In this work, we introduced a novel framework that integrates a LoRA (E. J. Hu et al., 2021) of pretrained diffusion and a fine-tuned multi-view diffusion model to ensure direct, stable, and realistic optimization of 3D-GS from random initialization without requiring additional priors for initialization.

## 3. Preliminaries

**Score Distillation Sampling (SDS)**. SDS (Poole et al., 2022) introduces a novel optimization framework that utilizes a pretrained text-to-image diffusion model to optimize 3D representations for text-to-3D generation. In this approach, a pretrained diffusion priors $p_t(x_t|y)$ with a noise prediction network $\epsilon_\phi(x_t, t, y)$, where $\phi$ represents the model parameters and $y$ is the input text prompt, is employed to guide the optimization of a 3D representation $\theta \in \Theta$, with $\Theta$ representing the space of all possible 3D representations. The goal of SDS is to align the 3D representation with the input text prompt $y$. Given a camera parameter $v$ and a differentiable rendering function $g(\cdot, v)$, define $q_0^\theta(x_0|v)$ as the distribution of $x_0 = g(\theta, v)$. Furthermore, $q_t^\theta(x_t|v)$ denotes the forward diffusion distribution of $q_0^\theta(x_0|v)$ at timestep $t$ in SDS. The optimization of $\theta$ is formulated as

$$\min_{\theta \in \Theta} \mathcal{L}_{SDS}(\theta) \coloneqq \mathbb{E}_{t,v}\left[\frac{\sigma_t}{\alpha_t}\omega(t)D_{KL}\big(q_t^\theta(x_t|v)\|p_t(x_t|y^v)\big)\right] \tag{1}$$

where $t \sim \mathcal{U}(0.02, 0.98)$, $\sigma_t, \alpha_t > 0$ are hyperparameters, $x_t = \alpha_t x_0 + \sigma_t \epsilon$, $\epsilon \sim \mathcal{N}(0, I)$, $\omega(t)$ is a weighting function dependent on the timestep $t$, and $y^v$ is the text prompt with view information $v$. Then, the gradient of SDS is approximated as

$$\nabla_\theta \mathcal{L}_{SDS}(\theta) \approx \mathbb{E}_{t,\epsilon,v}\left[\omega(t)\big(\hat{\epsilon}_\phi(x_t, t, y^c) - \epsilon\big)\frac{\partial x_0}{\partial \theta}\right] \tag{2}$$

where $\hat{\epsilon}_\phi(x_t, t, y^c) \coloneqq (1 + s)\epsilon_\phi(x_t, t, y^v) - s\epsilon_\phi(x_t, t, \emptyset)$ represents the CFG (Ho & Salimans, 2022), $\emptyset$ is a special "empty" text prompt used for the unconditional case, and $s > 0$ is the guidance scale. In this process, SDS is optimized independently for each viewpoint, while the pretrained diffusion model's lack of viewpoint priors leads to the Janus problem.

**3D Gaussian Splatting (3D-GS).** 3D-GS (Kerbl et al., 2023) is a groundbreaking method for novel view synthesis and 3D reconstruction, offering high-fidelity visual quality and efficient rendering. Unlike implicit 3D representation, 3D-GS models objects or scenes as a collection of anisotropic Gaussian distributions, each characterized by its center position $\boldsymbol{\mu} \in \mathbb{R}^3$, covariance matrix $\boldsymbol{\Sigma}$, color $\boldsymbol{c} \in \mathbb{R}^3$, and opacity $\alpha \in \mathbb{R}$. Each Gaussian is defined as follows

$$G(\boldsymbol{l}) = e^{-\frac{1}{2}(\boldsymbol{l})^T \Sigma^{-1}(\boldsymbol{l})} \tag{3}$$

where $\boldsymbol{l}$ is centered at the center position $\boldsymbol{\mu}$, and the covariance matrix $\boldsymbol{\Sigma}$ is parameterized by a scaling factor $\boldsymbol{s} \in \mathbb{R}^3$ and a rotation quaternion $\boldsymbol{q} \in \mathbb{R}^4$. The optimizable parameters of 3D-GS are thus defined as $\theta = \{\boldsymbol{\mu}_i, \boldsymbol{s}_i, \boldsymbol{q}_i, \boldsymbol{c}_i, \alpha_i\}$. During the rendering process, these 3D Gaussians are projected onto the image plane using a neural point-based method (Kopanas et al., 2021). The color $\mathcal{C}$ of each pixel is computed as

$$\mathcal{C}(\boldsymbol{r}) = \sum_{i \in \mathcal{P}} \boldsymbol{c}_i \sigma_i \prod_{j=1}^{i-1}(1 - \sigma_j), \sigma_i = \alpha_i G(\boldsymbol{l}_i) \tag{4}$$

where $\mathcal{P}$ denotes the ordered set of points overlapping the pixel on the ray $\boldsymbol{r}$, $\boldsymbol{c}_i$ and $\alpha_i$ denote the color and opacity of the $i$-th Gaussian, and $\boldsymbol{l}_i$ is the distance between the point and center position $\boldsymbol{\mu}_i$.

## 4. Method

In this section, we introduce Coupled Score Distillation (CSD), a novel framework for text-to-3D generation that effectively addresses the Janus problem, as illustrated in Figure 2. By formulating the task as a multi-view joint optimization problem, we derive a novel gradient-based optimization rule, termed Coupled Score Distillation (Section 4.1). We further provide a theoretical analysis to show that CSD effectively mitigates the Janus issue while preserving diversity, in contrast to single-view optimization (SDS (Poole et al., 2022)/VSD (Wang et al., 2023)) and fine-tuned multi-view approaches (Shi et al., 2023) (Section 4.2). Finally, we present the overall framework to achieve stable, photorealistic 3D-GS generation and high-quality mesh refinement (Section 4.3).

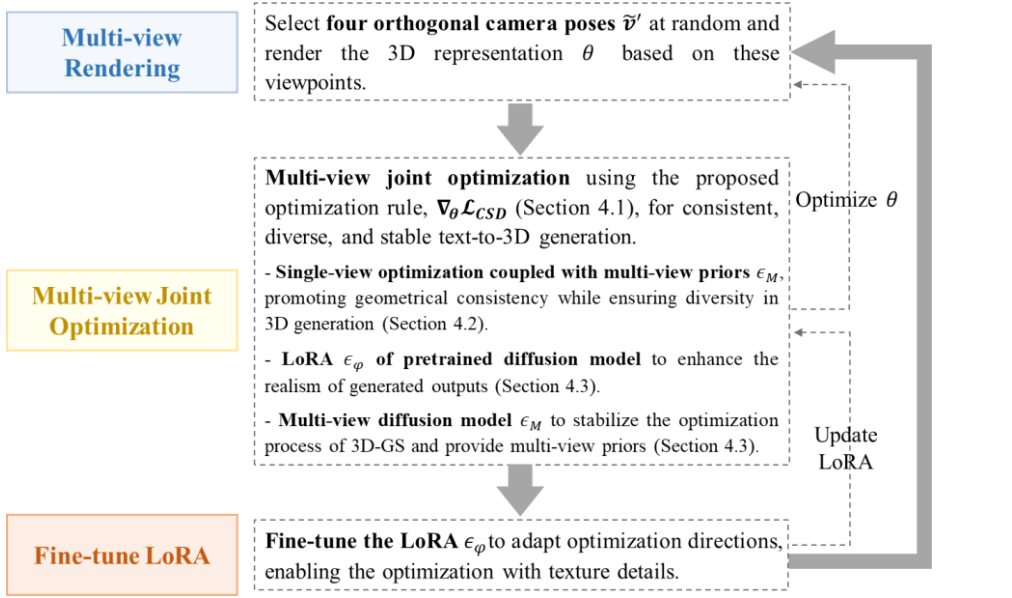

(a) Optimization process of the proposed CSD.

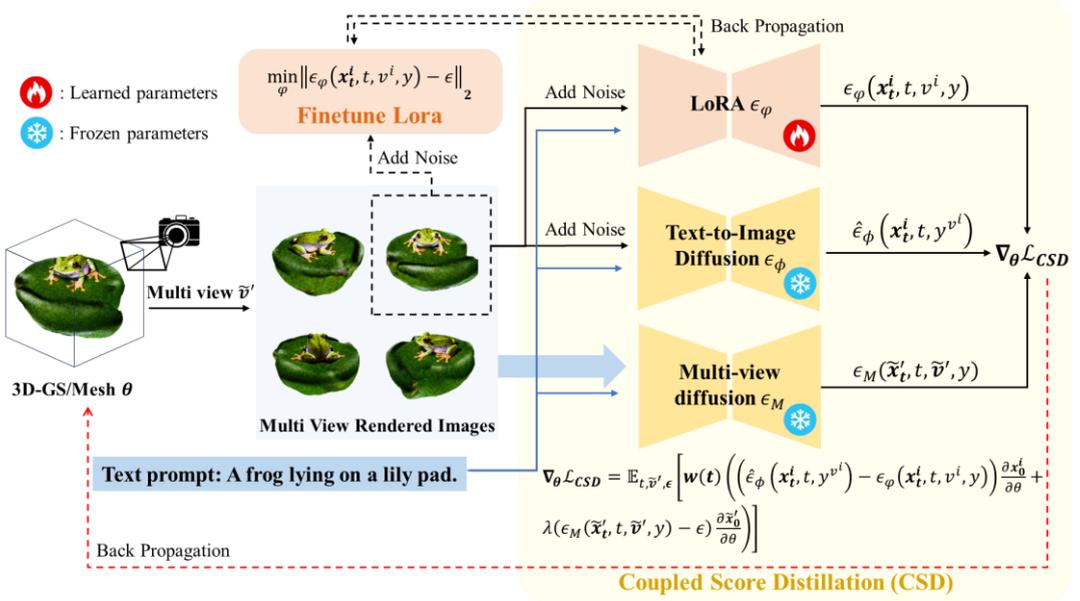

(b) Visualization of the proposed CSD.

Fig. 2. Overview of CSD. The 3D rendering is differentiable across four mutually orthogonal views $\tilde{v}'$. One of the orthogonal views is sent to the pretrained text-to-image diffusion model and LoRA, while all views are processed by the multi-view diffusion model to compute the CSD gradient. LoRA is fine-tuned based on the single rendered image.

## 4.1. Optimization Rule of Coupled Score Distillation

Theoretically, given a text prompt $y$, there exists a probability distribution over 3D representations that align with the semantic content of $y$ (Wang et al., 2023). Parameterizing the 3D representation (e.g., 3D-GS) as $\theta \in \Theta$, the distribution of the 3D parameter $\theta$ matching $y$ is denoted as $\mu(\theta|y)$. However, directly approximating $\mu(\theta|y)$ is hard due to the absence of paired text-3D data. To fix this issue, optimization-based methods leverage pretrained diffusion models to iteratively refine $\theta$, ensuring that a randomly initialized $\theta$ confirms $\mu(\theta|y)$ after optimization. To enforce consistency across multiple viewpoints, we model $\theta$ in terms of the multi-view joint distribution $q_0(\widetilde{x}_0|\widetilde{v}, y)$, where $\widetilde{x}_0 \coloneqq g(\theta, \widetilde{v})$ represents the multi-view renderings produced by the differentiable rendering function $g(\cdot, \widetilde{v})$ for the camera set $\widetilde{v} = \{v^0, v^1, \dots\}$. Meanwhile, $p_0(\widetilde{x}_0|\widetilde{v}, y)$ denotes the joint distribution of multi-view renderings from a sample within the distribution $\mu(\theta|y)$ that satisfies multi-view consistency. We align the rendered distribution $q_0(\widetilde{x}_0|\widetilde{v}, y)$ with $p_0(\widetilde{x}_0|\widetilde{v}, y)$ by solving:

$$\min_\theta \mathcal{L}_{CSD} = \min_\theta D_{KL}(q_0(\widetilde{x}_0|\widetilde{v}, y) \| p_0(\widetilde{x}_0|\widetilde{v}, y)) \tag{5}$$

It can be reformulated as a series of optimization problems, each associated with progressively diffused distributions indexed by timestep $t$ in alignment with the diffusion process of the diffusion model. Consequently, the solution to this problem is expressed as

$$\theta^* \coloneqq \arg\min_\theta \mathbb{E}_{t,\widetilde{v}} \left[ \frac{\sigma_t}{\alpha_t} \omega(t) D_{KL}(q_t(\widetilde{x}_t|\widetilde{v}, y) \| p_t(\widetilde{x}_t|\widetilde{v}, y)) \right] \tag{6}$$

where $p_t(\widetilde{x}_t|\widetilde{v}, y) \coloneqq p_t(x_t^0, x_t^1, \dots | v^0, v^1, \dots, y)$ represents the noisy joint distributions at timestep $t$ defined by the Gaussian transitions $p_t(\widetilde{x}_t|\widetilde{x}_0) = \mathcal{N}(\widetilde{x}_t|\alpha\widetilde{x}_0, \sigma_t^2 I)$, $\sigma_t, \alpha_t > 0$ are hyperparameters, and $\omega(t)$ is a time-dependent weighting function. Analogously, $q_t(\widetilde{x}_t|\widetilde{v}, y) \coloneqq q_t(x_t^0, x_t^1, \dots | v^0, v^1, \dots, y)$ is constructed in a similar manner.

Equation 6 cannot be directly optimized through training or finetuning a multi-view generative diffusion model due to the limited availability of text-3D paired data, which leads to overfitting. Therefore, we reformulate the optimization process using the probability product rule, expressed as $q_t(\widetilde{x}_t|\widetilde{v}, y) = q_t(\widetilde{x}_t|\widetilde{v}, y, x_t^i) \cdot q_t(x_t^i|v^i, y)$ and $p_t(\widetilde{x}_t|\widetilde{v}, y) = p_t(\widetilde{x}_t|\widetilde{v}, y, x_t^i) \cdot p_t(x_t^i|v^i, y)$, transforming the multi-view joint optimization into a single-view optimization task coupled with multi-view priors to ensure diversity and geometric consistency. The proposed approach is termed Coupled Score Distillation (CSD) and is formulated as

$$\nabla_\theta \mathcal{L}_{CSD} \triangleq \mathbb{E}_{t,\widetilde{v}} \left[ \sigma_t \omega(t) \left( \sum_{j=0}^{3} \left( \nabla_{x_t^{i+90j}} \log q_t(x_t^{i+90j}|v^{i+90j}, y) - \nabla_{x_t^{i+90j}} \log p_t(x_t^{i+90j}|v^{i+90j}, y) \right) \frac{\partial x_0^{i+90j}}{\partial \theta} \right. \right.$$
$$\left. \left. + \left( \nabla_{\widetilde{x}_t} \log q_t(\widetilde{x}_t|\widetilde{v}, y, \widetilde{x}_t') - \nabla_{\widetilde{x}_t} \log p_t(\widetilde{x}_t|\widetilde{v}, y, \widetilde{x}_t') \right) \frac{\partial \widetilde{x}_0}{\partial \theta} \right) \right] \tag{7}$$

where $\widetilde{x}_t' = (x_t^i, x_t^{i+90}, x_t^{i+180}, x_t^{i+270})$ represents four mutually orthogonal views from the cameras $\widetilde{v}' = (v^i, v^{i+90}, v^{i+180}, v^{i+270})$. The derivation can be found in Appendix A. Given the substantial number of iterations during the optimization process, where thousands of iterative steps involve repeated random sampling of viewpoints, we find that simplifying the sum of the four single-view losses to a single-view loss preserves optimization quality while significantly improving efficiency. Thus, the first single-view term is simplified by considering only one viewpoint per iteration. The resulting optimization is expressed as

$$\nabla_\theta \mathcal{L}_{CSD} \triangleq \mathbb{E}_{t,\tilde{v}} \left[ \omega(t) \left( \left( \underbrace{\left(-\sigma_t \nabla_{x_t^i} \log p_t(x_t^i|v^i,y)\right)}_{\text{score of single-view real images}} - \underbrace{\left(-\sigma_t \nabla_{x_t^i} \log q_t(x_t^i|v^i,y)\right)}_{\text{score of single-view rendered images}} \right) \frac{\partial x_0^i}{\partial \theta} \right. \right.$$
$$\left. \left. + \left( \underbrace{\left(-\sigma_t \nabla_{\tilde{x}_t} \log p_t(\tilde{x}_t|\tilde{v},y,\tilde{x}_t')\right)}_{\text{score of multi-view real images}} - \underbrace{\left(-\sigma_t \nabla_{\tilde{x}_t} \log q_t(\tilde{x}_t|\tilde{v},y,\tilde{x}_t')\right)}_{\text{score of multi-view rendered images}} \right) \frac{\partial \tilde{x}_0}{\partial \theta} \right) \right] \quad (8)$$

Equation 8 involves four score functions. The score of single-view real images $-\sigma_t \nabla_{x_t^i} \log p_t(x_t^i|v^i,y)$ and the score of multi-view real images $-\sigma_t \nabla_{\tilde{x}_t} \log p_t(\tilde{x}_t|\tilde{v},y,\tilde{x}_t')$ can be approximated by a pretrained text-to-image diffusion model with CFG $\hat{\epsilon}_\phi\left(x_t^i, t, y^{v^i}\right)$ and a fine-tuned multi-view diffusion model $\epsilon_M(\tilde{x}_t', t, \tilde{v}', y)$, respectively. The score of multi-view rendered images $-\sigma_t \nabla_{\tilde{x}_t} \log q_t(\tilde{x}_t|\tilde{v},y,\tilde{x}_t')$ is represented by the added noise $\epsilon \sim \mathcal{N}(0, I)$ during the diffusion process. Furthermore, inspired by ProlificDreamer (Wang et al., 2023), the score of single-view rendered images $-\sigma_t \nabla_{x_t^i} \log q_t(x_t^i|v^i,y)$ is estimated by a LoRA of diffusion model $\epsilon_\varphi(x_t^i, t, v^i, y)$ conditioned on camera parameter $v^i$ and text prompt $y$, which leverages a common CFG weight for visually refined 3D generation. The LoRA is fine-tuned on the rendered single-view image with diffusion objective during optimization process

$$\min_\varphi \mathbb{E}_{t,\epsilon,v^i} \left[ \left\| \epsilon_\varphi(x_t^i, t, v^i, y) - \epsilon \right\|_2^2 \right] \quad (9)$$

where $x_t^i = \alpha_t g(\theta, v^i) + \alpha_t \epsilon$. Finally, the gradient-based optimization rule is

$$\nabla_\theta \mathcal{L}_{CSD} \triangleq \mathbb{E}_{t,\tilde{v}',\epsilon} \left[ \omega(t) \left( \left( \hat{\epsilon}_\phi(x_t^i, t, y^{v^i}) - \epsilon_\varphi(x_t^i, t, v^i, y) \right) \frac{\partial x_0^i}{\partial \theta} + \lambda (\epsilon_M(\tilde{x}_t', t, \tilde{v}', y) - \epsilon) \frac{\partial \tilde{x}_0'}{\partial \theta} \right) \right] \quad (10)$$

where $y^{v^i}$ represents the text prompt with view information $v^i$, and $\lambda$ serves as a regularization coefficient to balance the discrepancies between the limited viewpoints generated by $\epsilon_M(\tilde{x}_t', t, \tilde{v}', y)$ and the joint viewpoints $\tilde{x}_t$. The detailed pseudocode of the iterative training process is provided in Algorithm 1.

Algorithm 1. Pseudocode for the iterative training process of CSD

---
**Coupled Score Distillation**: Adam optimizer (Kingma & Ba, 2014) with $(\beta_1, \beta_2)$, $eps = 1e - 15$ for 3D representation $\theta$; AdamW (Loshchilov & Hutter, 2017) optimizer with $(\beta_1, \beta_2) = (0.9, 0.999)$, $esp = 1e - 8$ for the LoRA parameter $\varphi$.

---
**Input**: Text-to-image diffusion model $\hat{\epsilon}_\phi$; multi-view diffusion model $\epsilon_M$; frequency $k$ for learning the LoRA; an input text prompt $y$; learning rate $\eta_1$ and $\eta_2$ for 3D representation $\theta$ and LoRA parameter $\varphi$
**Initialize**: The 3D representation $\theta$ and LoRA $\epsilon_\varphi$ parameterized by $\varphi$.
**for** $iter = 1$ **to** $iter_{total}$ **do**
    Randomly sample four mutually orthogonal camera poses $\tilde{v}' = (v^i, v^{i+90}, v^{i+180}, v^{i+270})$
    Render the 3D representation $\theta$ at poses $\tilde{v}'$ to get four 2D images $\tilde{x}_0' = (x_0^i, x_0^{i+90}, x_0^{i+180}, x_0^{i+270})$
    $\theta \leftarrow \theta - \eta_1 \mathbb{E}_{t,\tilde{v}',\epsilon} \left[ \omega(t) \left( \left( \hat{\epsilon}_\phi(x_t^i, t, y^{v^i}) - \epsilon_\varphi(x_t^i, t, v^i, y) \right) \frac{\partial x_0^i}{\partial \theta} + \lambda (\epsilon_M(\tilde{x}_t', t, \tilde{v}', y) - \epsilon) \frac{\partial \tilde{x}_0'}{\partial \theta} \right) \right]$
    **if** $iter \% k == 0$ **then**
        $\varphi \leftarrow \varphi - \eta_2 \mathbb{E}_{t,\epsilon,v^i} \left[ \left\| \epsilon_\varphi(x_t^i, t, v^i, y) - \epsilon \right\|_2^2 \right]$
    **end if**
**end for**
\# $k = 1, (\beta_1, \beta_2) = (0.9, 0.99)$ for optimizing 3D-GS and $k = 10, (\beta_1, \beta_2) = (0.9, 0.999)$ for refining mesh texture

*4.2. Single-view Optimization with Multi-view Priors for Diversity and Consistency*

In this section, we compare the proposed CSD with single-view optimization, such as SDS/VSD (Poole et al., 2022; Wang et al., 2023) and fine-tuned multi-view approaches, such as MVDream (Shi et al., 2023), to demonstrate its effectiveness, as illustrated in Figure 3.

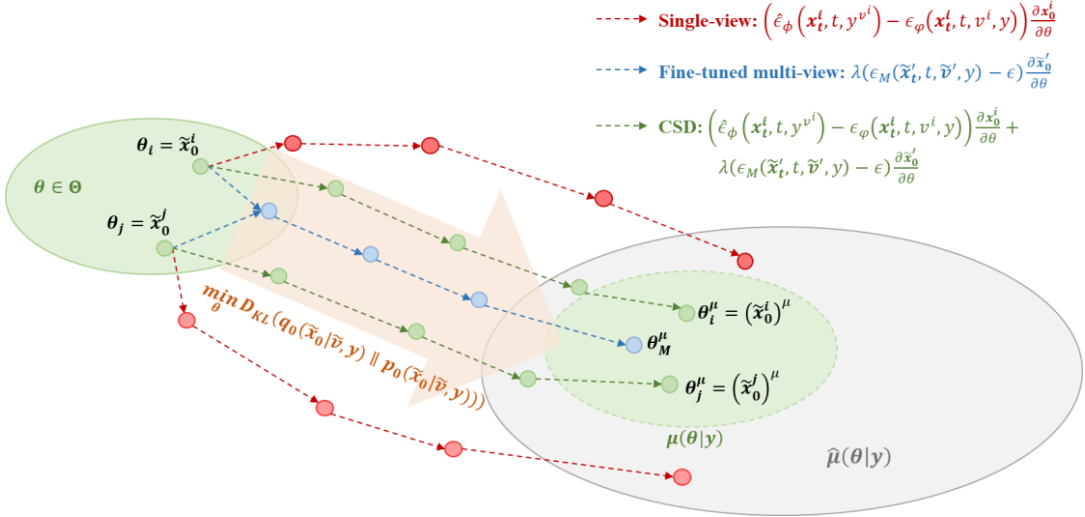

Fig. 3. Comparison of the proposed CSD with single-view and fine-tuned multi-view optimization methods. $\hat{\mu}(\theta|y)$ represents the multi-view join distribution, including both geometrically consistent and inconsistent samples. $\mu(\theta|y)$ denotes the distribution of geometrically consistent samples. The light red arrow area represents the method of multi-view joint optimization. The dashed line indicates the update direction of parameter $\theta$ guided by the diffusion model.

**The single-view optimization methods** leverage the generation diversity of pretrained text-to-image diffusion models, where varying 3D initializations produce diverse 2D images to guide optimization, enabling diverse 3D assets generation to align with the same text prompt $y$ (red dashed line in Figure 3). However, the absence of viewpoint information in the diffusion model and the lack of inter-viewpoint correlation during optimization result in generated 3D assets that follow $\hat{\mu}(\theta|y)$ (grey region in Figure 3), being prone to the Janus problem.

**The fine-tuned multi-view methods** refine the text-to-image diffusion model with viewpoint information and multi-view renderings to establish correlations across viewpoints. However, due to the limited size of the 3D dataset, the model tends to overfit, producing identical multi-view images for different 3D initialization parameters during optimization. This overfitting leads to highly similar 3D outputs, reducing the diversity of the generated results (blue dashed line in Figure 3).

**The proposed CSD framework** enables multi-view joint optimization by integrating single-view optimization with fine-tuned multi-view prior constraints, ensuring geometrically consistent gradient directions (green dashed line in Figure 3). This approach achieves a synergistic balance between the diversity offered by single-view optimization and the geometric consistency imposed by multi-view finetuned methods, effectively approximating the target distribution (represented by the green subregion within the grey area in Figure 3). The efficacy of the proposed CSD in generating diverse and geometrically consistent outcomes is further validated through experiments detailed in

Section 5.1.

*4.3. Framework for Stable and Realistic Optimization*

Building on the proposed CSD optimization rule, we present a unified framework for stable, diverse, and high-quality text-to-3D generation, compassing guidance diffusion models and optimization strategies for 3D-GS and mesh. More training details are provided in Appendix B.

**Guidance Diffusion Models**. The optimized gradient direction in the proposed CSD framework integrates two key components: single-view optimization with LoRA adaptation to promote diversity and refine visual details, and multi-view priors to ensure geometric consistency and stability. For single-view optimization, we adopt SD2.1-base (Rombach et al., 2022; *Stable-Diffusion-2-1-Base*, n.d.) to approximate $\hat{\epsilon}_\phi\left(x_t^i, t, y^{v^i}\right)$ and SD2.1-v (Rombach et al., 2022; *Stable-Diffusion-2-1*, n.d.) with a LoRA module (E. J. Hu et al., 2021) to approximate $\epsilon_\varphi(x_t^i, t, v^i, y)$ and gradient directions for capturing finer visual details by fine-tuning LoRA during the optimization process while preserving diversity. Nonetheless, in the initial stages of 3D-GS optimization, LoRA introduces noise harmful to stability. Based on the proposed CSD, MVDream (Shi et al., 2023) is employed as multi-view priors to impose cross-view constraints, enabling multi-view joint optimization and enhancing the stability of the optimization process.

**Optimization Strategy**. The framework adopts a two-stage approach (C.-H. Lin et al., 2023; Wang et al., 2023) to optimize 3D-GS and mesh for generating photorealistic and geometrically consistent results. In the first stage, high-resolution 3D-GS (e.g., 1024) is optimized starting from random positions within a unit sphere, utilizing an annealed time scheduler (Wang et al., 2023) and progressively scaling the resolution from 64 to 1024. Throughout this process, densification and pruning are applied intermittently without resetting opacity. In the second stage, a texture mesh, represented by DMTet (T. Shen et al., 2021), is extracted from the optimized 3D-GS using a local density query (Tang, Ren, et al., 2024). The extracted mesh is then fine-tuned by independently optimizing its geometry and texture to achieve photorealistic details, following the Fantasia3D (R. Chen et al., 2023).

## 5. Experiments

In this section, we evaluate the proposed approach through comprehensive experiments. We benchmark the proposed CSD against state-of-the-art methods for text-to-3D generation, demonstrating its capability to produce photorealistic, semantically, and geometrically consistent 3D assets while preserving the diversity of generated results. Additionally, ablation studies are conducted to evaluate the contribution of individual components within the proposed CSD.

*5.1. Text-to-3D Generation*

We present the results generated by the proposed CSD in Figure 1, showcasing high-fidelity 3D-GS and mesh outputs, with additional examples provided in Appendix D. Furthermore, we quantitatively and qualitatively evaluate the method's performance in terms of geometrically and semantically consistent generation, efficient and robust optimization of 3D-GS, and diversity for general text-to-3D generation against baseline approaches.

**Geometrically and Semantically Consistent Generation.** We quantitatively and qualitatively compare our method with nine 3D generation baselines, including seven optimization-based methods (Magic3D (C.-H. Lin et al., 2023), ProlificDreamer (Wang et al., 2023), Score Distillation via Reparametrized DDIM (SDI) (Lukoianov et al., 2024), DreamGaussian (Tang, Ren, et al., 2024), GaussianDreamer (Yi et al., 2024), GSGEN (Z. Chen et al., 2024), and LucidDreamer (Liang et al., 2024)), a text-to-image-to-3D method (LGM (Tang, Chen, et al., 2024)), and a feed-forward method (Trellis (Xiang et al., 2024)). As Magic3D is not publicly available, we re-implement it using the threestudio (Y.-T. Liu et al., 2023) project, an open-source and unified framework for 3D content generation, to ensure a consistent and fair comparison. We provide quantitative and qualitative comparisons in the following.

*Quantitative Results.* To comprehensively assess the semantic and geometric consistency of text-to-3D methods, we adopt a benchmark of 30 text prompts, following the evaluation protocol of DreamControl (Huang et al., 2024) (see Appendix C for the prompt list). Geometric consistency is quantified by the Janus Rate (JR), which is defined as the proportion of results exhibiting unreasonable geometry. Semantic consistency is measured using the CLIPScore (CS) (Hessel et al., 2022), which evaluates the alignment between the generated 3D content and the corresponding text prompt. Details of the evaluation metrics are provided in Appendix C. As shown in Table 1, our CSD consistently outperforms other methods across both metrics. Additionally, the experimental results further confirm that, while recent feed-forward methods (e.g., Trellis) have improved generation speed and view consistency, they still lag behind optimization-based approaches (e.g., DreamControl, SDI, GaussianDreamer, GSGEN, LucidDreamer, and the proposed CSD) in generation quality, due to the limited scale and cartoon style of current 3D training datasets. In comparison, our CSD couples single-view optimization with multi-view priors for multi-view joint optimization, leading to more precise geometric consistency, enhanced texture quality, and improved semantic alignment with text prompts.

Table 1. Quantitative results. The proposed CSD outperforms all baseline methods in both geometric and semantic consistency.

| Method | JR (%) ↓ | CS (%) ↑ |
|---|---|---|
| DreamFusion* (Poole et al., 2022) | 36.67 | 26.36 |
| Magic3D* (C.-H. Lin et al., 2023) | 53.33 | 26.59 |
| ProlificDreamer* (Wang et al., 2023) | 56.67 | 26.69 |
| DreamControl* (Huang et al., 2024) | 10.00 | 28.14 |
| SDI (Lukoianov et al., 2024) | 53.33 | 28.14 |
| DreamGaussian (Tang, Ren, et al., 2024) with MVDream (Shi et al., 2023) | 10.00 | 22.17 |
| DreamGaussian (Tang, Ren, et al., 2024) with SDS | 66.66 | 22.33 |
| GaussianDreamer (Yi et al., 2024) | 26.67 | 29.23 |
| GSGEN (Z. Chen et al., 2024) | 76.67 | 28.31 |
| LucidDreamer (Liang et al., 2024) | 33.33 | 27.89 |
| LGM (Tang, Chen, et al., 2024) | 23.33 | 29.42 |
| Trellis (Xiang et al., 2024) | 10.00 | 27.42 |
| **CSD (Ours)** | **3.33** | **30.23** |

* represents the results reported by DreamControl (Huang et al., 2024).

*Qualitative Results.* To visually compare generation quality, we select six text prompts and display multi-view renderings of each result to qualitatively evaluate geometric and visual fidelity, as illustrated in Figure 4. Experimental results demonstrate that other optimization-based approaches frequently exhibit the Janus problem—e.g., frogs or corgis with multiple faces, or Iron Man with additional limbs—stemming from unresolved perspective biases in 2D diffusion models. Text-to-image-to-3D methods heavily depend on the quality of intermediate images, while feed-forward models often generate cartoonish textures due to limitations in existing 3D datasets. In contrast, our CSD addresses the inherent perspective bias of 2D diffusion models by incorporating multi-view priors and effectively exploits the high-fidelity texture from 2D diffusion models to generate geometrically consistent and realistic 3D assets.

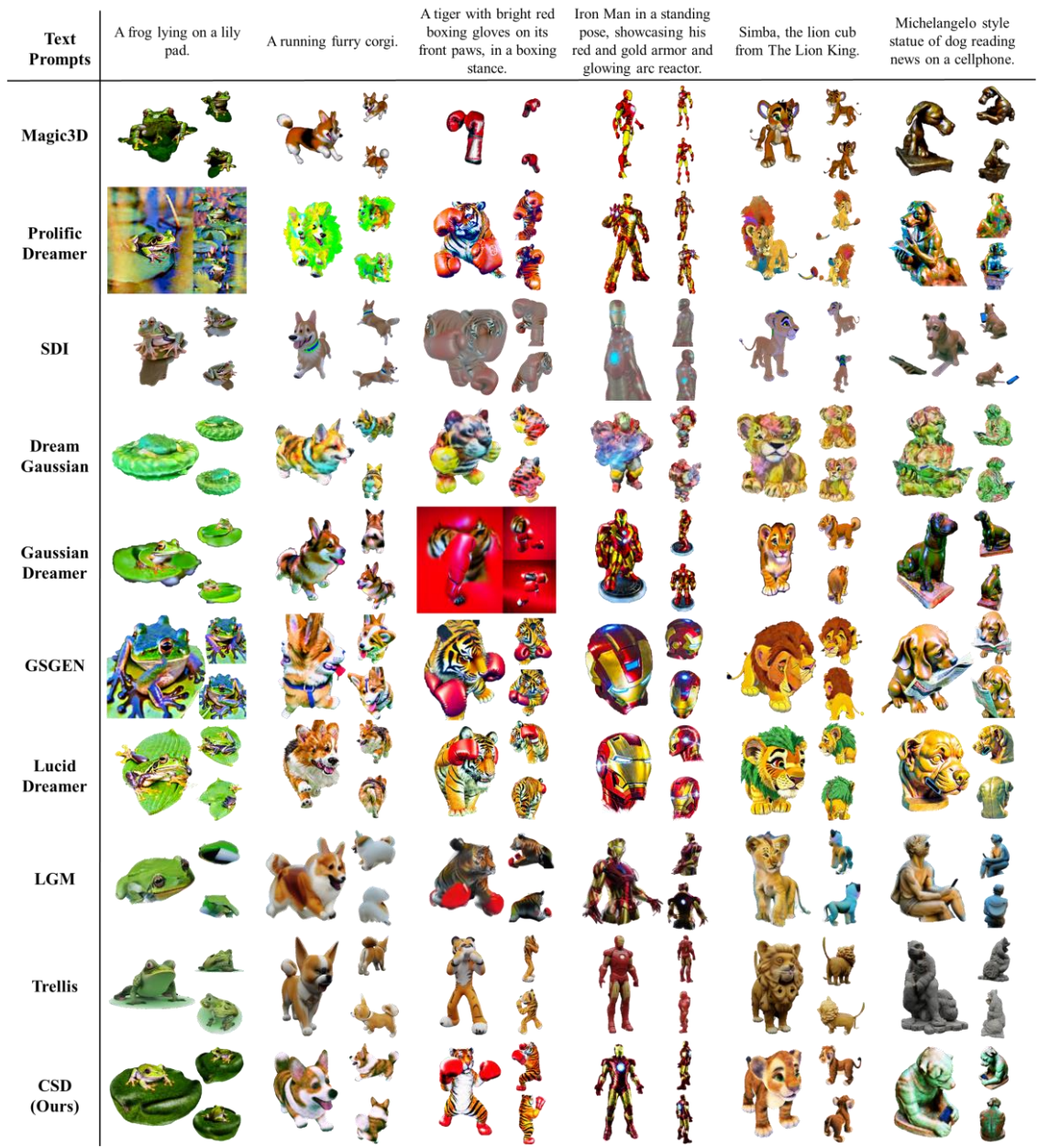

Fig. 4. Qualitative comparison of geometric consistency between the proposed CSD framework and baselines, including

Magic3D (C.-H. Lin et al., 2023), ProlificDreamer (Wang et al., 2023), SDI (Lukoianov et al., 2024), DreamGaussian (Tang, Ren, et al., 2024) with MVDream (Shi et al., 2023), GSGEN (Z. Chen et al., 2024), LucidDreamer (Liang et al., 2024)), LGM (Tang, Chen, et al., 2024), and Trellis (Xiang et al., 2024). Results for Magic3D are reproduced using the open-source threestudio (Y.-T. Liu et al., 2023). Each object is represented by three images, captured from horizontal viewpoints separated by 120°.

**Efficient and Robust Optimization of 3D-GS**. To evaluate the robustness and efficiency of the proposed framework for optimizing 3D-GS, we compare the proposed CSD method from random initialization against GaussianDreamer (Yi et al., 2024), which employs point cloud priors (Jun & Nichol, 2023) for initialization, and GSGEN (Z. Chen et al., 2024), which applies point cloud prior (Nichol et al., 2022) for both initialization and 3D loss regularization in geometry optimization. As shown in Figure 5, while both GaussianDreamer and GSGEN produce detailed textures, they are still prone to the multi-face problem and generate inconsistent geometries. In contrast, the proposed CSD framework, starting from random initialization, achieves geometrically consistent and realistic results, effectively mitigating the Janus problem. Furthermore, by initializing from the generated 3D-GS and integrating the two-stage pipeline, we obtain high-quality 3D meshes with reasonable geometry and refined textures.

**Diversity Generation.** To evaluate the diversity of generated results, we compare the proposed CSD with the fine-tuned multi-view diffusion model MVDream (Shi et al., 2023), which leverages NeRF as the 3D representation optimized via SDS. To examine potential overfitting, we randomly initialize the 3D representation using different seeds and optimize it accordingly, as illustrated in Figure 6. The results show that the proposed CSD generates diverse 3D assets from the same text prompt, such as Spiderman in various dancing poses, whereas MVDream produces nearly identical outputs. This highlights that our method effectively approximates a distribution aligned with the text prompt, while the fine-tuned approach is limited to a single sample from the distribution due to overfitting.

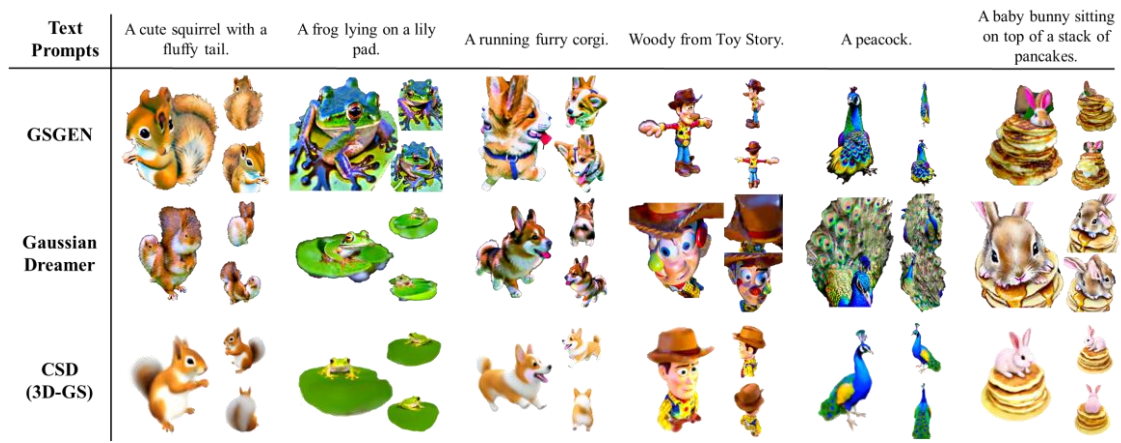

Fig. 5. Comparison of efficient and robust generation between the proposed CSD framework and state-of-the-art optimization methods of 3D Gaussian Splatting, including GaussianDreamer (Yi et al., 2024) and GSGEN (Z. Chen et al., 2024). GaussianDreamer uses Shap-E (Jun & Nichol, 2023) for initialization, GSGEN applies Point-E (Nichol et al., 2022) for initialization and regularization, while the proposed CSD framework achieves consistent and realistic results from random initialization. Each object is represented by three images, captured from horizontal viewpoints separated by

120°.

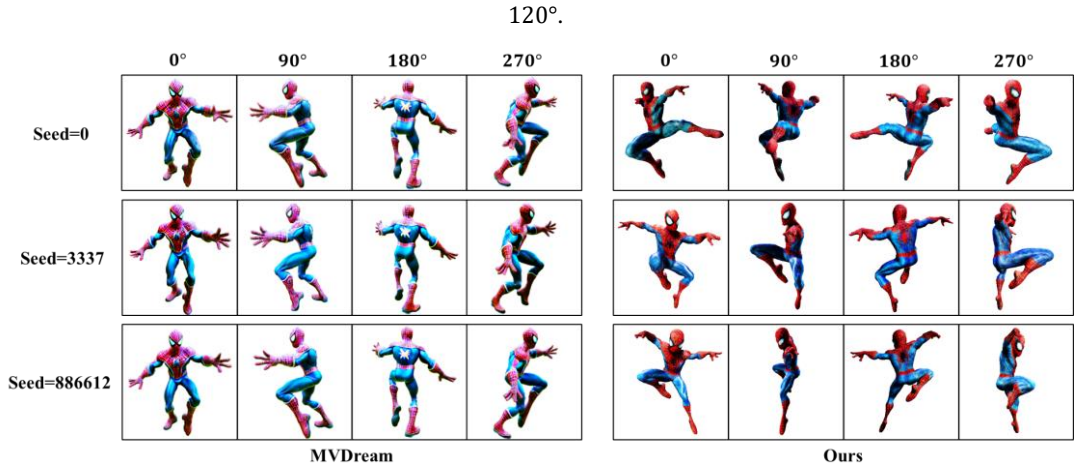

Text Prompt: **A DSLR photo of a Spiderman dancing, Marvel character, highly detailed 3D model.**

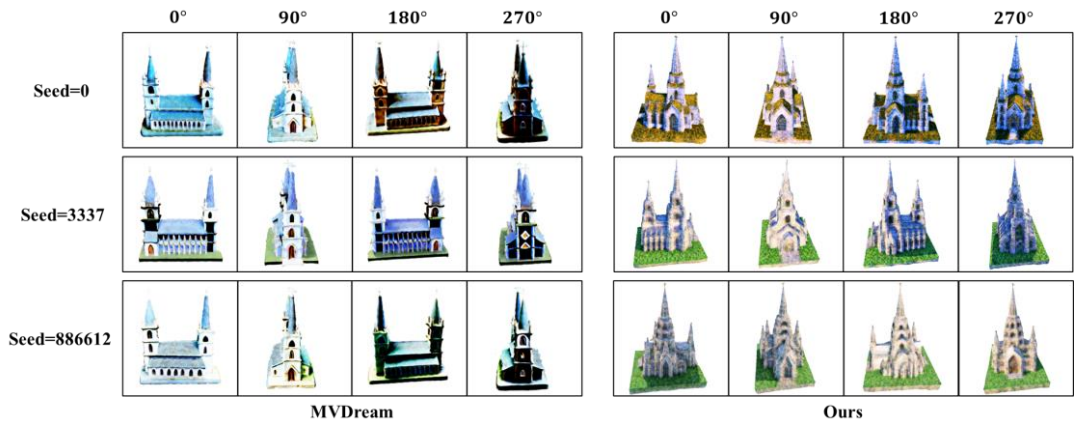

Text Prompt: **A church with towering spires and intricate details, 3D**

Fig. 6. Comparison of diversity generation between the proposed CSD and fine-tuned multi-view methods. Results are obtained by optimizing different seed initializations using the proposed CSD and MVDream (Shi et al., 2023), respectively.

## 5.2. Ablation Study

In this section, we perform an ablation study to analyse the contribution of each component in the proposed CSD framework. Specifically, we investigate the impact of the LoRA module and the multi-view prior constraint (e.g., MVDream (Shi et al., 2023)) in the optimization process of 3D-GS, evaluating their roles independently, as illustrated in Figure 7.

**LoRA Module**. To assess the impact of the LoRA module, we replace it with the added noise $\epsilon$ from diffusion process and compare the generated results with and without LoRA. The results, as shown in the third and fourth columns of Figure 7, demonstrate that employing a learnable LoRA of pretrained diffusion model to approximate diffuse noise significantly enhances the detail and realism of the optimized 3D outputs. By fine-tuning LoRA during optimization, gradient directions are effectively adapted to capture finer visual details, mitigating the over-smoothing artifacts observed in the generated 3D results.

**Multi-view Prior Constraint**. To enhance the geometric consistency of generated 3D assets, the framework incorporates MVDream as a multi-view prior constraint. Its effectiveness is validated by removing the multi-view joint optimization term and comparing the results with those produced by the full CSD framework, as illustrated in the second and fourth columns of Figure 7. The results demonstrate that multi-view constraints significantly improve geometric coherence while stabilizing the optimization process of 3D-GS.

By combining the LoRA module and multi-view priors, the proposed framework facilitates direct and stable optimization of 3D-GS from random initialization, effectively generating diverse and high-quality 3D assets while resolving the multi-face Janus problem.

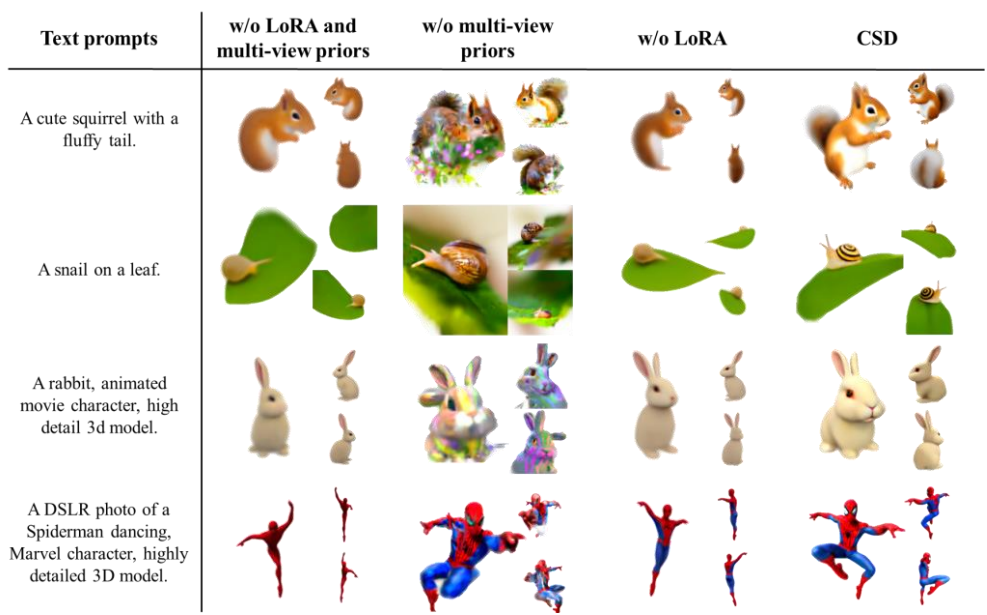

Fig. 7. Ablation study results illustrating the impact of each component in the optimization process of 3D Gaussian Splatting using the proposed CSD framework. Each object is represented by three images captured from horizontal viewpoints, spaced 120° apart.

## 6. Conclusion

In this paper, we proposed Coupled Score Distillation (CSD), a framework that couples multi-view joint distribution priors to address multi-face artifacts while ensuring the stable optimization of 3D Gaussian Splatting (3D-GS) with random initialization and high-reality 3D mesh for text-to-3D generation. By reformulating the optimization process as a multi-view joint optimization problem, we derived the Coupled Score Distillation that integrates single-view optimization with multi-view priors to achieve coherent 3D generation. We further introduced a framework that integrates a LoRA of diffusion model and a fine-tuned multi-view diffusion model, enabling stable and efficient 3D generation with 3D-GS. Additionally, we presented a two-stage method for geometrically consistent and realistic 3D-GS and mesh generation based on the proposed CSD. Experimental results demonstrate the efficiency and competitive quality of our approach, yielding diverse, photorealistic, and geometrically consistent 3D outputs.

## 7. Limitation and Future Work

Although the proposed CSD effectively generates geometrically consistent and high-quality 3D objects, it struggles with producing assets that fall outside the scope of the diffusion model's training data. This issue can be mitigated by increasing underrepresented data samples or fine-tuning the model with out-of-range data. Additionally, although CSD can produce high-fidelity 3D assets with strong geometric consistency, the generation process can take up to two hours. In future work, we plan to leverage CSD's high-quality generation capabilities to train a feed-forward network, aiming to achieve fast and efficient text-to-3D synthesis.

**Impact Statement**

This paper presents work whose goal is to advance the field of text-to-3D Generation. While generative models can substantially improve creative efficiency, they may disrupt traditional labour markets by automating human-driven tasks. Moreover, like other generative technologies, our method carries the risk of misuse for producing misleading or harmful content. Ensuring responsible use and implementing proactive measures are essential to mitigate these risks and promote ethical deployment.

**Declaration of Interests**

The authors declare that they have no known competing financial interests or personal relationships that could have appeared to influence the work reported in this paper.

**Acknowledgments**

This work was supported by the National Key R&D Program of China (Grant No. 2023YFC3805700), National Natural Science Foundation of China (Grant No. 52408171), and China Post-doctoral Science Foundation (Grant No. 2024M764196).

**Appendix A Theory of Coupled Score Distillation**

Equation 6 leverages the properties of the diffusion model to reformulate the variational inference problem in Equation 5 into a series of optimization tasks, each aligning two diffused distributions indexed by $t$. This reformulation simplifies the process, as the diffused distributions are closer to a standard Gaussian. Furthermore, Equation 6 is derived as

$$\begin{aligned}
\nabla_\theta \mathcal{L}_{CSD} &\triangleq \nabla_\theta \mathbb{E}_{t,\tilde{v}}\left[\frac{\sigma_t}{\alpha_t}\omega(t)D_{KL}(q_t(\tilde{x}_t|\tilde{v},y)\|p_t(\tilde{x}_t|\tilde{v},y))\right] \\
&\triangleq \mathbb{E}_{t,\tilde{v}}\left[\frac{\sigma_t}{\alpha_t}\omega(t)\left(\nabla_{\tilde{x}_t}\log q_t(\tilde{x}_t|\tilde{v},y) - \nabla_{\tilde{x}_t}\log p_t(\tilde{x}_t|\tilde{v},y)\right)\frac{\partial \tilde{x}_t}{\partial \tilde{x}_0}\cdot\frac{\partial \tilde{x}_0}{\partial \theta}\right] \\
&\triangleq \mathbb{E}_{t,\tilde{v}}\left[\sigma_t\omega(t)\left(\nabla_{\tilde{x}_t}\log q_t(\tilde{x}_t|\tilde{v},y) - \nabla_{\tilde{x}_t}\log p_t(\tilde{x}_t|\tilde{v},y)\right)\frac{\partial \tilde{x}_0}{\partial \theta}\right]
\end{aligned} \quad (A.1)$$

where $p_t(\tilde{x}_t|\tilde{x}_0) = \mathcal{N}(\tilde{x}_t|\alpha\tilde{x}_0, \sigma_t^2 I)$. To ensure the multi-view consistency and diversity of optimization results, we reformulate the multi-view joint optimization problem as a single-view optimization coupled with multi-view prior constraints, utilizing the diversity generation capability of the pretrained text-to-image diffusion model with the multi-view priors of the fine-tuned multi-view diffusion model and achieving diverse optimization results without the multi-face Janus problem. Based on the probability product rule, expressed as $q_t(\tilde{x}_t|\tilde{v},y) = q_t(\tilde{x}_t|\tilde{v},y,x_t^i)\cdot q_t(x_t^i|v^i,y)$ and $p_t(\tilde{x}_t|\tilde{v},y) = p_t(\tilde{x}_t|\tilde{v},y,x_t^i)\cdot p_t(x_t^i|v^i,y)$, Equation (A.1) can be rewritten as

$$\begin{aligned}
\nabla_\theta \mathcal{L}_{CSD} &\triangleq \mathbb{E}_{t,\tilde{v}}\Big[\sigma_t\omega(t)\Big(\nabla_{\tilde{x}_t}\log\big(q_t(\tilde{x}_t|\tilde{v},y,x_t^i)\cdot q_t(x_t^i|v^i,y)\big) \\
&\qquad - \nabla_{\tilde{x}_t}\log\big(p_t(\tilde{x}_t|\tilde{v},y,x_t^i)\cdot p_t(x_t^i|v^i,y)\big)\Big)\frac{\partial \tilde{x}_0}{\partial \theta}\Big] \\
&\triangleq \mathbb{E}_{t,\tilde{v}}\Big[\sigma_t\omega(t)\Big(\big(\nabla_{x_t^i}\log q_t(x_t^i|v^i,y) - \nabla_{x_t^i}\log p_t(x_t^i|v^i,y)\big)\frac{\partial x_0^i}{\partial \theta} \\
&\qquad + \big(\nabla_{\tilde{x}_t}\log q_t(\tilde{x}_t|\tilde{v},y,x_t^i) - \nabla_{\tilde{x}_t}\log p_t(\tilde{x}_t|\tilde{v},y,x_t^i)\big)\frac{\partial \tilde{x}_0}{\partial \theta}\Big)\Big] \triangleq \cdots \\
&\triangleq \mathbb{E}_{t,\tilde{v}}\Big[\sigma_t\omega(t)\Big(\sum_{j=0}^{3}\big(\nabla_{x_t^{i+90j}}\log q_t(x_t^{i+90j}|v^{i+90j},y) \\
&\qquad - \nabla_{x_t^{i+90j}}\log p_t(x_t^{i+90j}|v^{i+90j},y)\big)\frac{\partial x_0^{i+90j}}{\partial \theta} \\
&\qquad + \big(\nabla_{\tilde{x}_t}\log q_t(\tilde{x}_t|\tilde{v},y,\tilde{x}_t') - \nabla_{\tilde{x}_t}\log p_t(\tilde{x}_t|\tilde{v},y,\tilde{x}_t')\big)\frac{\partial \tilde{x}_0}{\partial \theta}\Big)\Big]
\end{aligned} \quad (A.2)$$

where $\tilde{x}_t' = \left(x_t^i, x_t^{i+90}, x_t^{i+180}, x_t^{i+270}\right)$ is a significantly smaller subset of $\tilde{x}_t$. In this work, $x_t^i, x_t^{i+90}, x_t^{i+180}, x_t^{i+270}$ correspond to four mutually orthogonal rendered images obtained from the camera poses $\tilde{v}' = (v^i, v^{i+90}, v^{i+180}, v^{i+270})$.

## Appendix B  More Details on Implementation and Hyper-Parameters

**3D Gaussian Splatting details**. The 3D Gaussians are initialized with an opacity of 0.1 and a neutral grey color, distributed within a sphere of radius 0.5. During optimization, the rendering resolution is progressively increased, starting at 128 and scaling to 256, 512, and 1024 at 10%, 30%, and 50% of the total iterations, respectively. The background color is randomly selected as either white or black. Initialization begins with 1000 random particles, followed by iterative densification and pruning every 250 iterations until 1500 iterations, without resetting opacity. For densification and pruning, a gradient threshold of 0.01 is used, and Gaussians with an opacity below 0.01 or a maximum scaling factor exceeding 0.05 are removed.

**Mesh details.** The geometry initialization of mesh is performed using local density queries (Tang, Ren, et al., 2024) with an empirical threshold of 0.2 to extract voxels, followed by computing the Signed Distance Function (SDF) using *distance_transform_edt* from SciPy (Virtanen et al., 2020) to initial DMTet (T. Shen et al., 2021). During geometry finetuning, both the SDF and the deformation parameters of DMTet are treated as learnable variables and jointly optimized. For texture fine-tuning, a hash-grid positional encoding (Müller et al., 2022) and an MLP are utilized to predict the mesh texture.

**Guidance details.** All diffusion models employed for guidance in this study are implemented using the Hugging Face Diffusers library (von Platen et al., 2022/2024). For single-view guidance, we utilize the *stabilityai/stablediffusion−2−1−base* as the single-view prior and use v-prediction (Salimans & Ho, 2022) to train the LoRA of diffusion model with *stabilityai/stablediffusion−2−1*. In the LoRA implementation, the camera pose $v^i$ is processed by a two-layer MLP and incorporated into the timestep embedding at each U-Net block, following the approach in ProlificDreamer (Wang et al., 2023). For multi-view guidance, we adopt the *MVDream/MVDream* as the multi-view priors, ensuring consistent optimization across different perspectives.

**Training details of 3D-GS.** The optimization of 3D Gaussian Splatting (3D-GS) is conducted $4k$ iterations, employing an annealed time scheduler that transitions from $t \sim \mathcal{U}(0.02, 0.98)$ to $t \sim \mathcal{U}(0.02, 0.50)$ after $2k$ iterations. For the learning rate of 3D-GS, we adopt different learning rates for distinct parameters. The position learning rate decays from $1 \times 10^{-3}$ to $2 \times 10^{-5}$ within the first 1500 iterations, while fixed learning rates are assigned as follows: 0.01 for feature, 0.05 for opacity, $5 \times 10^{-3}$ for scaling, and $1 \times 10^{-3}$ for rotation. Optimization is carried out using the Adam optimizer (Kingma & Ba, 2014). For the regularization coefficient λ in the proposed CSD, values are set dynamically within the range of 0.1 to 1.0, tailored to the specific characteristics of the target object during 3D-GS optimization.

**Training details of mesh.** The mesh is refined by separately optimizing its geometry and texture to achieve photorealistic detail, similar to the approach in Fantasia3D (R. Chen et al., 2023). Geometry is first fine-tuned using a normal map and SDS (Poole et al., 2022) with a CFG of 100. The texture is then refined under the supervision of CSD, with a CFG of 7.5. The geometry fine-tuning process consists of $15k$ iterations, with an annealed time scheduler that transitions from $t \sim \mathcal{U}(0.02, 0.98)$ to $t \sim \mathcal{U}(0.02, 0.50)$ after $5k$ iterations, and a learning rate of $5 \times 10^{-3}$ optimized using Adam optimizer (Kingma & Ba, 2014). The mesh fine-tuning stage consists of $4k$ iterations, beginning with a warm-up phase during the first $1k$ iterations to initialize the hash encoder and MLP, with an annealed time scheduler that transitions from $t\sim\mathcal{U}(0.02,0.98)$ to $t\sim\mathcal{U}(0.02,0.50)$ after $2k$ iterations. The hash encoder uses a learning rate of 0.05, while the MLP is optimized with a learning rate of $5\times10^{-3}$, both using the Adam optimizer. The regularization coefficient $\lambda$ for the proposed CSD is fixed at 0.1 in the fine-tuned stage of texture. Additionally, the rendering resolution is fixed at 512 for both geometry and texture fine-tuning.

**Other training details.** Camera poses are sampled within a radius range of $[2.0, 2.5]$, with a vertical field of view (FOV) in $[40°, 70°]$, azimuth angles in $[-180°, 180°]$, and elevation angles in $[-90°, 30°]$. The 3D-GS optimization stage takes approximately 50 minutes, while the mesh optimization stage requires about 1.5 hours, both executed on two NVIDIA 3090 GPUs with a batch size of 1.

## Appendix C 30 Text Prompts and Evaluation Metrics

Here are the 30 text prompts used for the evaluations:

1. A chimpanzee dressed like Henry VIII king of England.
2. A white astronaut is riding a brown horse.
3. Michelangelo style statue of dog reading news on a cellphone.
4. A blue jay standing on a large basket of rainbow macarons.
5. A baby bunny sitting on top of a stack of pancakes.
6. A statue of angel.
7. Batman is riding a moto.
8. Elon musk, using a laptop.
9. Lionel Messi in a suit, holding the Ballon d'Or.
10. A wide angle zoomed out DSLR photo of Tower Bridge made out of gingerbread and candy.
11. A highly detailed sand castle.
12. A model of a house in Tudor style.
13. A pavilion in a Chinese garden.
14. Woody from Toy Story.
15. A standing Captain America, Marvel character.
16. Spider Man.
17. A gundam.
18. A wizard.
19. Astronaut.
20. A classic Packard car.
21. A vase with pink flowers.
22. A plate piled high with chocolate chip cookies.
23. A plate of fried chicken and waffles with maple syrup on them.

24. A corgi.
25. Simba, the lion cub from The Lion King.
26. A dragon-cat hybrid.
27. A peacock.
28. An elephant.
29. A teddy bear.
30. A Tesla Model3 sedan.

**Janus Rate (JR)**. To assess geometric consistency, the Janus Rate is defined as the proportion of generated results exhibiting the Janus problem. As illustrated in Figure C.1, following the DreamControl (Huang et al., 2024), a generation is identified as a Janus problem if it exhibits: (1) multi-face, multi-hand, multi-leg, or similar issues; (2) obvious content drift; (3) serious paper-thin generation. JR is computed as the ratio of such inconsistent cases to the total number of generated samples.

**CLIPScore (CS)**. The CLIPScore (Hessel et al., 2022) is computed based on the CLIP similarity between the input prompt and a single rendered view of the generated object. For each prompt $y$, an image $x^y$ is rendered from the generated 3D asset for each method, using a fixed camera pose at $-45°$ elevation and $30°$ azimuth. The final score is obtained by averaging the CLIP similarities across all prompts: $\frac{1}{|Y|}\sum_{y \in Y} S_{CLIP}(y, x^y)$.

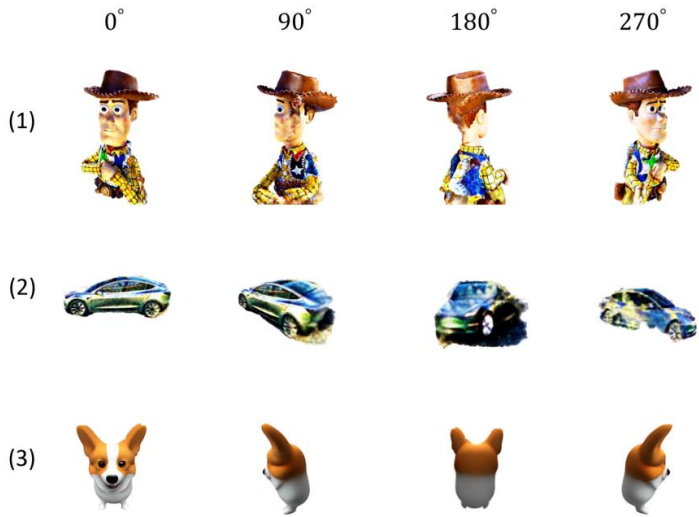

Fig. C.1. Examples of Janus problems provided by DreamControl (Huang et al., 2024), including: (1) multi-face, multi-hand, multi-leg, or similar issues; (2) obvious content drift; (3) serious paper-thin generation.

**Appendix D  Additional Experiment Results of 3D Gaussian Splatting and Textured Meshes**

Additional results showcasing 3D Gaussian Splatting and textured mesh generation using the proposed CSD framework are presented in Figures D.1 and D.2, respectively. For video demonstrations of the generated 3D results, please refer to the anonymous webpage: https://showresults.github.io/CSD/.

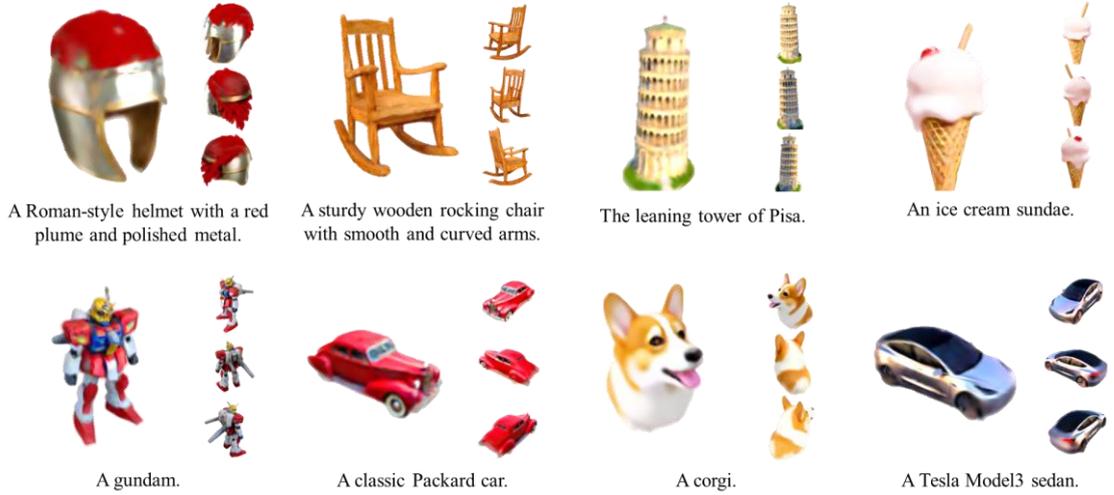

Fig. D.1. More generated results of 3D Gaussian Splatting with the proposed CSD. Each object is depicted with four images captured from horizontal viewpoints spaced 90° apart.

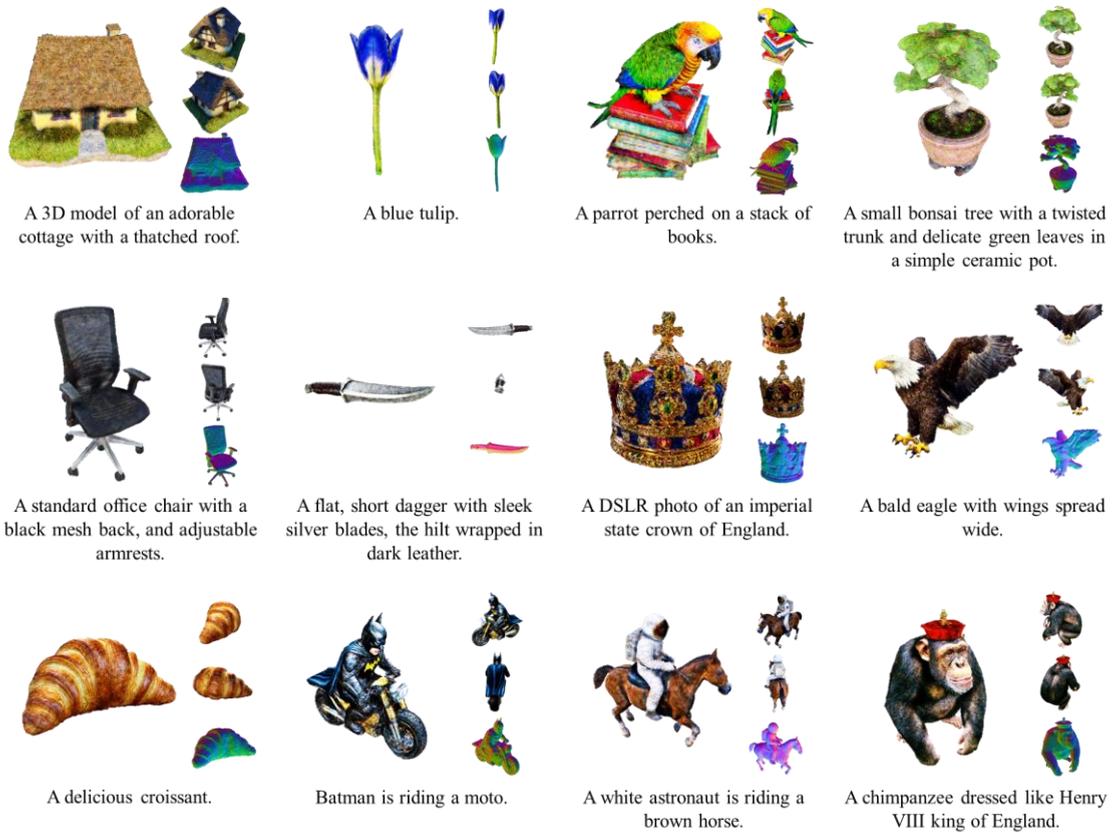

Fig. D.2. More generated results of textured mesh with the proposed CSD. Each object is depicted with three images captured from horizontal viewpoints spaced 120° apart, along with a corresponding mesh representation.